%% file: arxiv.tex
\documentclass[10pt,twocolumn,letterpaper]{article}

\usepackage[pagenumbers]{cvpr} 

\renewcommand{\thefootnote}{\*}
\definecolor{cvprblue}{rgb}{0.21,0.49,0.74}
\usepackage[pagebackref,breaklinks,colorlinks,allcolors=cvprblue]{hyperref}
\usepackage{tabularx}
\usepackage{multirow}
\usepackage[table]{xcolor}
\newcolumntype{Y}{>{\centering\arraybackslash}X}

\def\paperID{*****} 
\def\confName{CVPR}
\def\confYear{2026}

\title{ProPhy: Progressive Physical Alignment for Dynamic World Simulation}
\author{
Zijun Wang\textsuperscript{1,2}\footnotemark[1], 
Panwen Hu\textsuperscript{3}\footnotemark[1], 
Jing Wang\textsuperscript{1}, 
Terry Jingchen Zhang\textsuperscript{4}, 
Yuhao Cheng\textsuperscript{5}, \\ 
Long Chen\textsuperscript{5}, 
Yiqiang Yan\textsuperscript{5}, 
Zutao Jiang\textsuperscript{2}, 
Hanhui Li\textsuperscript{1}\footnotemark[2], 
Xiaodan Liang\textsuperscript{1,2,3}\footnotemark[2] \\[0.5em]
\textsuperscript{1}Shenzhen Campus of Sun Yat-sen University, 
\textsuperscript{2}Peng Cheng Laboratory \\
\textsuperscript{3}Mohamed bin Zayed University of Artificial Intelligence, 
\textsuperscript{4}ETH Zürich, 
\textsuperscript{5}Lenovo Research \\
}
\begin{document}

\twocolumn[{
\renewcommand\twocolumn[1][]{#1}%
\maketitle
\vspace{-12mm}
\begin{center}
    \centering
    \includegraphics[width=0.95\linewidth]{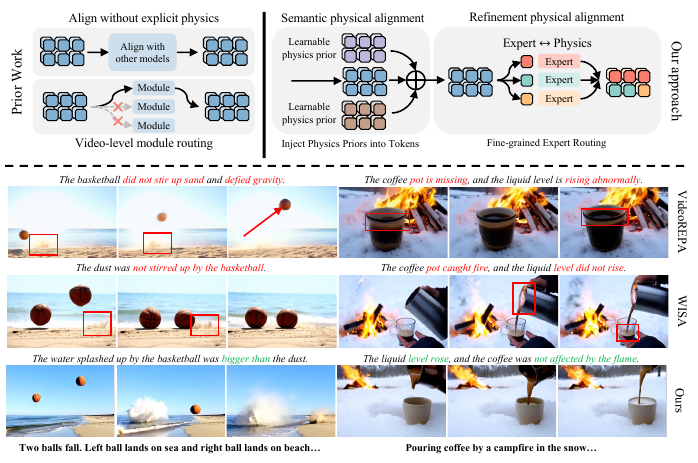}
    \vspace{-12pt}
\captionof{figure}{
\textit{Top-left}: Prior work relies on \textbf{implicit alignment} without explicitly modeling physical priors, or adopts \textbf{video-level routing} that assigns whole videos to coarse experts without fine-grained physical reasoning.
\textit{Top-right}: Overview of \textbf{ProPhy}, a \textbf{progressive physical alignment framework} that explicitly injects learnable physics priors and progressively refines them via token-level expert routing, allowing different experts to specialize in distinct physical phenomena.
\textit{Bottom}: Qualitative comparison in complex scenarios. Red boxes and arrows highlight violations of physical laws in prior methods.
}
    \label{fig:teaser}
\end{center}
}]
\renewcommand{\thefootnote}{\fnsymbol{footnote}}
\footnotetext[1]{Equal contribution}
\footnotetext[2]{Corresponding authors}
\renewcommand{\thefootnote}{\arabic{footnote}}
\input{sec/0_abstract}    
\input{sec/1_intro}
\input{sec/2_related}
\input{sec/3_method}
\input{sec/4_exp}
\input{sec/5_conc}

\section*{Acknowledgement}
This work is supported by Scientific Research Innovation Capability Support Project for Young Faculty (No.ZYGXQNJSKYCXNLZCXM-I28), National Natural Science Foundation of China (NSFC) under Grants No.62476293 and No.62372482, Major Key Project of PCL (Grant No. PCL2025A17), and General Embodied AI Center of Sun Yat-sen University.

{
    \small
    \bibliographystyle{ieeenat_fullname}
    \bibliography{main}
}


\end{document}

%% file: sec/0_abstract.tex
\begin{abstract}
Recent advances in video generation have shown remarkable potential for constructing world simulators. However, current models still struggle to produce physically consistent results, particularly when handling large-scale or complex dynamics. This limitation arises primarily because existing approaches respond isotropically to physical prompts and neglect the fine-grained alignment between generated content and localized physical cues. To address these challenges, we propose ProPhy, a Progressive Physical Alignment Framework that enables explicit physics-aware conditioning and anisotropic generation. ProPhy employs a two-stage Mixture-of-Physics-Experts mechanism for discriminative physical prior extraction, where Semantic Experts infer semantic-level physical principles from textual descriptions, and Refinement Experts capture token-level physical dynamics. This mechanism allows the model to learn fine-grained, physics-aware video representations that better reflect underlying physical laws. Furthermore, we introduce a physical alignment strategy that transfers the physical reasoning capabilities of vision-language models into the Refinement Experts, facilitating a more accurate representation of dynamic physical phenomena. Extensive experiments on physics-aware video generation benchmarks demonstrate that ProPhy produces more realistic, dynamic, and physically coherent results than existing state-of-the-art methods. Project Page: \url{https://zijunwa.github.io/prophy/}
\end{abstract}

%% file: sec/1_intro.tex
\section{Introduction}
\label{sec:intro}

Video generation~\cite{cogvideox, wan, hunyuan} has recently emerged as a powerful paradigm for visual content synthesis, driving a wide range of applications such as creative media production~\cite{hu2024storyagent, wang2025dreamvideo, jiangVACEAllinOneVideo2025a} and robotic simulation~\cite{cosmos, bruceGenieGenerativeInteractive2024a, zhouRoboDreamerLearningCompositional2024}. Despite the rapid progress in visual realism, current video generation models still struggle to produce \textit{physically plausible} results, which limits their ability to function as world simulators~\cite{brooks2024video, zhu2024sora} capable of reproducing realistic physical dynamics~\cite{lin2025exploringevolutionphysicscognition}.

Recently, several studies~\cite{wisa, videorepa, ji2025physmaster, li2025pisa, chen2025hierarchical} have explored enhancing the physical plausibility of generated videos. For instance, VideoREPA~\cite{videorepa} employs a distillation strategy to transfer physical knowledge from foundation models. Other researchers focus on alignment-based optimization; PhysMaster~\cite{ji2025physmaster} and PISA~\cite{li2025pisa} utilize reinforcement learning and reward modeling to induce physical behaviors, while PhysHPO~\cite{chen2025hierarchical} introduces a hierarchical preference alignment framework to optimize video consistency across multiple granularities. However, as shown in the top row of Figure~\ref{fig:teaser}, these approaches primarily rely on internalizing physical priors during training, lacking explicit physical guidance during inference. This often leads to generations that deviate from fundamental physical laws in complex scenarios. In contrast, WISA~\cite{wisa} introduces explicit guidance by identifying physical categories from prompts and employing a Mixture-of-Physics-Experts (MoPE) structure. Nevertheless, as illustrated in the bottom row of Figure~\ref{fig:teaser}, WISA’s guidance is largely global-level; it struggles to capture fine-grained processes when multiple physical phenomena coexist or occur within localized regions.

To address the aforementioned limitations, we identify two key challenges that must be overcome: (a) \textbf{Explicit physical guidance}, which enables the representations of different physical laws to become more discriminative, thereby capturing distinct physical characteristics; and (b) \textbf{Fine-grained physical alignment}, which allows different spatial regions within a video to respond accurately to localized physical cues.  To this end, we propose \textbf{ProPhy}, a Progressive Physical Alignment Framework that not only adaptively extracts physics-specific priors from textual descriptions but also refines and injects these priors into the spatial regions corresponding to particular physical phenomena, achieving physically consistent and spatially anisotropic video generation. Specifically, unlike previous methods that employ a uniform architecture or a single-level physical prior to model all physical laws (top-left of Figure~\ref{fig:teaser}), we introduce a two-stage MoPE mechanism in our framework, which mainly consists of a \textbf{Semantic Expert Block} and a \textbf{Refinement Expert Block}. These two modules progressively extract and refine hierarchical physical priors (top-right of Figure~\ref{fig:teaser}). The Semantic Expert Block first infers physics-specific priors directly from textual prompts, which are subsequently fused with visual features and further processed by the Refinement Expert Block to learn token-wise physical priors. This design facilitates the formation of more discriminative and physically expressive representations.

Furthermore, to enhance the regional response to underlying physical laws, we propose an innovative physical alignment strategy that guides the refinement experts during training to generate spatially anisotropic representations. At the current stage, vision-language models (VLMs)~\cite{bai2025qwen2} exhibit a more detailed and reliable localization of physical phenomena compared to generative models. Building on this observation, our strategy aligns the spatial distributions predicted by the physics experts with those inferred from the VLMs, thereby transferring its fine-grained physical localization into the generative process. By reinforcing the model’s sensitivity to localized physical priors, our method achieves more accurate and stable generation of dynamic physical processes, particularly in high-motion scenarios.

We evaluate our approach on the physics-related video generation benchmark~\cite{bansalVideoPhy2ChallengingActionCentric2025}, and the experimental results demonstrate that ProPhy consistently outperforms state-of-the-art methods across both standard and challenging scenarios, particularly in maintaining dynamic physical consistency. Our main contributions can be summarized as follows: 1) we propose a two-stage MoPE design for physical prior extraction, enabling the model to capture physics-specific priors and learn discriminative physical representations more precisely; 2) Moreover, an innovative fine-grained physical alignment strategy is introduced to allow the Refinement Expert to learn fine-grained physical priors from VLMs, thereby improving the prediction of spatially physical distributions; 3) Finally, we conduct extensive experiments across multiple backbone architectures to validate the effectiveness and generality of our framework, as well as the importance of fine-grained physical modeling.

%% file: sec/2_related.tex
\section{Related Work}
\label{sec:related}

\subsection{Video Generation as World Simulator}
The advent of diffusion models~\cite{ho2020denoising, song2020denoising, lipman2022flow} has significantly advanced the field of visual generation, enabling generators to produce highly realistic images~\cite{rombachHighResolutionImageSynthesis2022b, esser2024scaling, flux2024}. In particular, Diffusion Transformers~\cite{peebles2023scalable, maLatteLatentDiffusion2025} (DiT) leverage the scaling capabilities of transformers within diffusion models, and when combined with large model and dataset scales, have led to a series of impressive video generation models~\cite{wan, kling2024, cogvideox, hunyuan, brooks2024video, cosmos} capable of producing realistic videos. Among these, Sora~\cite{brooks2024video} explores the use of extensive video data to develop a general-purpose world simulator.

Despite these advances, current video generation models still struggle to fully capture the underlying physical principles of the real world, focusing primarily on the appearance of scenes and objects~\cite{kang2024far, liu2025generative, motamed2025generative}. Increasing model size or dataset scale alone does not enable generators to learn the physical laws embedded in scenes and textual descriptions. This gap between video generators and true world simulators results in videos that are visually convincing but often logically inconsistent, limiting their potential for accurate physics-aware generation.

\subsection{Physics-Aware Video Generation}
Recent research has explored enhancing the physical awareness of video generation models through various approaches. 1) Physics simulation methods~\cite{xie2025physanimator, montanaro2024motioncraft, zhang2024physdreamer} first predict object dynamics from images and then render them into videos. PhysGen~\cite{liu2024physgen} simulates rigid-body dynamics to estimate how objects respond to external forces, generating keyframes that are subsequently rendered into videos. Similarly, PhysMotion~\cite{tan2024physmotion} animates static scenes into dynamic videos using the Material Point Method~\cite{stomakhin2013material} (MPM) and refines details with a diffusion model. However, these methods require predefined dynamics parameters and explicit physical rules, limiting their generalization to diverse, unconstrained scenarios for true world simulation. 2) Learning-based approaches extract physical laws directly from video by leveraging temporal relationships. VideoREPA~\cite{videorepa}, for example, utilizes inter-frame visual feature relations extracted by a visual encoder to guide the diffusion model, producing videos that more closely reflect real-world dynamics. 3) Methods incorporating external physics priors extract structured physical information using large vision-language or language models to guide generation. PhysT2V~\cite{xue2025phyt2v} iteratively adjusts textual instructions with LLMs to correct physical violations in generated videos, while WISA~\cite{wisa} constructs structured physical representations—including descriptions, categories, and attributes—and integrates them into the generation process. NewtonGen~\cite{yuan2025newtongen} predicts physical states using Neural Newtonian Dynamics (NND), enabling physics-controllable video synthesis. 

Despite these advances, existing methods for extracting and leveraging physical information in general scenarios largely operate at the sample level. However, different physical phenomena within a scene often appear at distinct spatial locations. Sample-level guidance therefore lacks fine-grained alignment with localized physical cues, dispersing physical awareness and failing to focus on critical areas. In this work, we propose a framework that integrates spatially-specific physics priors, enabling enhanced physical perception while maintaining realistic motion dynamics.

%% file: sec/3_method.tex
\section{Method}
\label{sec:method}

\begin{figure*}
    \centering
    \includegraphics[width=\linewidth]{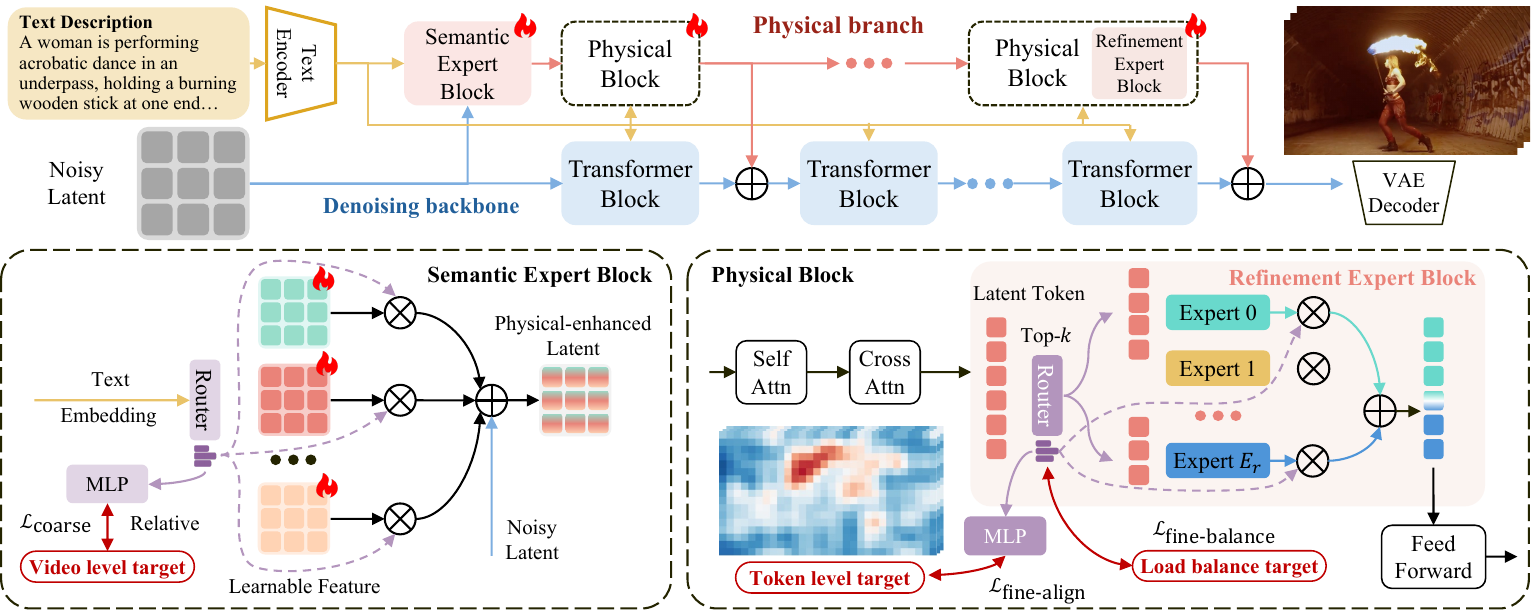}
    \vspace{-12pt}
    \caption{Overview of our proposed ProPhy framework. ProPhy uses a progressive physical alignment design, consisting of the \textbf{Semantic Expert Block} and the \textbf{Refinement Expert Block}. During inference, the model runs end-to-end and aligns physics categories through our proposed blocks.}
    \vspace{-12pt}
    \label{fig:framework}
\end{figure*}

\subsection{Overall Framework}

Our framework follows the paradigm of mainstream video diffusion models (VDMs)~\cite{cogvideox, wan} and aims to generate \textit{physically plausible} videos from text. Given a text prompt $I$, our end-to-end inference pipeline first analyzes the input and extracts the underlying physical priors. It then denoises an initial noise sample $\boldsymbol{X}_T$ into a clean video $\boldsymbol{X}_0$. Unlike prior methods~\cite{videorepa, ji2025physmaster, wisa} that rely solely on coarse, video-level physical constraints, our framework adopts a progressive physical alignment strategy. As shown in Figure~\ref{fig:framework}, ProPhy is built upon latent video diffusion backbones such as WAN~\cite{wan} and CogVideoX~\cite{cogvideox}. We further introduce a dedicated \textbf{Physical Branch}, which consists of a \textbf{Semantic Expert Block (SEB)}, multiple \textbf{Physical Blocks (PB)}, and a \textbf{Refinement Expert Block (REB)} attached to the final PB.


The SEB activates the relevant semantic physics experts based on the implicit physical cues in the text prompt, producing video-level physical priors. These priors are progressively refined by the PBs and further processed by the REB to obtain fine-grained, token-level physical priors. The refined priors are then injected into the backbone video representations, enabling spatially anisotropic responses to physical phenomena. To preserve the pretrained backbone’s semantic understanding and rendering ability, each PB adopts the same architecture as its corresponding transformer block and is initialized with its weights. The PB outputs are sequentially injected into the video latent, allowing the model to accumulate physical information in a progressive manner. The number of PBs is adapted to the depth of each backbone model and the available GPU memory during training. Detailed configurations are provided in the \textbf{Supplementary Material}.





\subsection{Progressive Physical Routing}

\paragraph{Semantic Expert Block}

The SEB operates at the video level. It contains a set of learnable physical basis maps, each representing a distinct aspect of physical knowledge, which together serve as physical experts to provide physics-aware priors for video representation. The contribution of each expert is determined by a semantic router, which identifies the underlying physical semantics from the input text prompt and dynamically selects the most relevant experts for activation.
Formally, we define $E_s$ learnable physical basis maps $\boldsymbol{B}_e \in \mathbb{R}^{N \times C}$ for $e = 1, \dots, E_s$. Each $\boldsymbol{B}_e$ shares the same shape as the backbone’s visual latent $\boldsymbol{X}_t \in \mathbb{R}^{N \times C}$. Here $N = (F/r_f) \times (H/r_s) \times (W/r_s)$ is the number of latent tokens. $(F, H, W)$ are the video length, height, and width. $(r_f, r_s)$ are temporal and spatial downsampling ratios. $C$ is the latent dimension. For a sample’s text embedding $y$, the semantic router outputs normalized weights $\boldsymbol{\rho}_p \in \mathbb{R}^{E_p}$ to control the contribution of each physical basis map. Thus, the physics-enhanced latent is represented as
\begin{equation}
    \tilde{\boldsymbol{X}} = \boldsymbol{X} + \sum_{e=1}^{E_s}\boldsymbol{\rho}_p^e \boldsymbol{B}_e.
\end{equation}


During training, small batch sizes will make the standard top-$k$ MoE~\cite{moe} prone to mode collapse~\cite{qiuDemonsDetailImplementing2025}, where only a few experts are repeatedly activated. To address this, we adopt the continuous weighted formulation described above. The resulting feature $\tilde{\boldsymbol{X}}$, serves as the global physical prior for subsequent refinement.


\paragraph{Refinement Expert Block}


Unlike the SEB, the REB operates at the token level. It also contains a set of experts, each implemented by a linear layer, together with a \textit{refinement router} that predicts the underlying physical law associated with each token and selects the most relevant top-$k$ experts to provide physics-aware priors for that token. Specifically, for each token $\tilde{\boldsymbol{x}} \in \mathbb{R}^C$ in the physics-enhanced latent, the refinement router outputs a weight vector $\boldsymbol{\rho}_r \in \mathbb{R}^{E_r}$ representing the probability distribution over the physical laws associated with this token, where $E_r$ is the number of the refinement experts. Because the number of tokens is large and the proposed fine-grained alignment strategy (discussed in the next section) is applied, the risk of mode collapse is considerably reduced in this stage. We therefore adopt a standard MoE strategy:
\begin{equation}
    \tilde{\boldsymbol{x}}' = \sum_{i \in \text{argtop}_k\boldsymbol{\rho}_r} \boldsymbol{\rho}_r^i \,\boldsymbol{e}_\theta^i(\tilde{\boldsymbol{x}}),
\end{equation}
where $\boldsymbol{e}_\theta^i$ denotes the forward function of the $i$-th expert. 


\subsection{Physical Alignment Objectives}


During training, each expert learns its physical knowledge under the guidance of its router. Our framework introduces two separate alignment objectives for the two routers. Experts at different levels therefore learn complementary aspects of physical reasoning. For example, in the SEB, some experts focus on combustion phenomena, and others capture reflection behaviors. We also apply a load-balancing loss to encourage fair expert activation. These components together yield a diverse pool of physics-aware experts and support the refinement process that follows.
\vspace{-6pt}
\paragraph{Semantic Alignment}

At this stage, we align the semantic experts’ routing weights $\boldsymbol{\rho}_p$ with global physical semantics. We use WISA-80K \cite{wisa} and its per-video physical category vector $\boldsymbol{q}_s \in \mathbb{R}^{E_{\text{wisa}}}$, where $E_{\text{wisa}}$ is the number of annotated categories. A linear layer maps $\boldsymbol{\rho}_s$ to the same dimension as $\boldsymbol{q}_s$. For each batch of size $B$, we compute a cosine-similarity-based pairwise matrix:
\begin{equation}
    \boldsymbol{P}_s^{i, j} = \frac{\boldsymbol{\rho}_s^{(i)} \cdot \boldsymbol{\rho}_s^{(j)}}{\|\boldsymbol{\rho}_s^{(i)}\| \,\|\boldsymbol{\rho}_s^{(j)}\|},
    \label{eq:3}
\end{equation}
where $\boldsymbol{\rho}_s^{(i)}$ is the vector for the $i$-th sample in the batch. Using the same procedure as Equation~\ref{eq:3}, we compute the label matrix $\boldsymbol{Q}_s \in \mathbb{R}^{B \times B}$. The semantic alignment objective is formulated as
\begin{equation}
    \mathcal{L}_{\text{coarse}} = \sum_{1 \leq i < j \leq B} \|\boldsymbol{P}_s^{i, j} - \boldsymbol{Q}_{s}^{i, j}\|_2.
\end{equation}

With this objective, samples within a batch that belong to the same physical category tend to share similar routing weights, whereas samples from different categories exhibit more divergent routing distributions.


\paragraph{Fine-grained Alignment}

\begin{figure}
    \centering
    \includegraphics[width=\columnwidth]{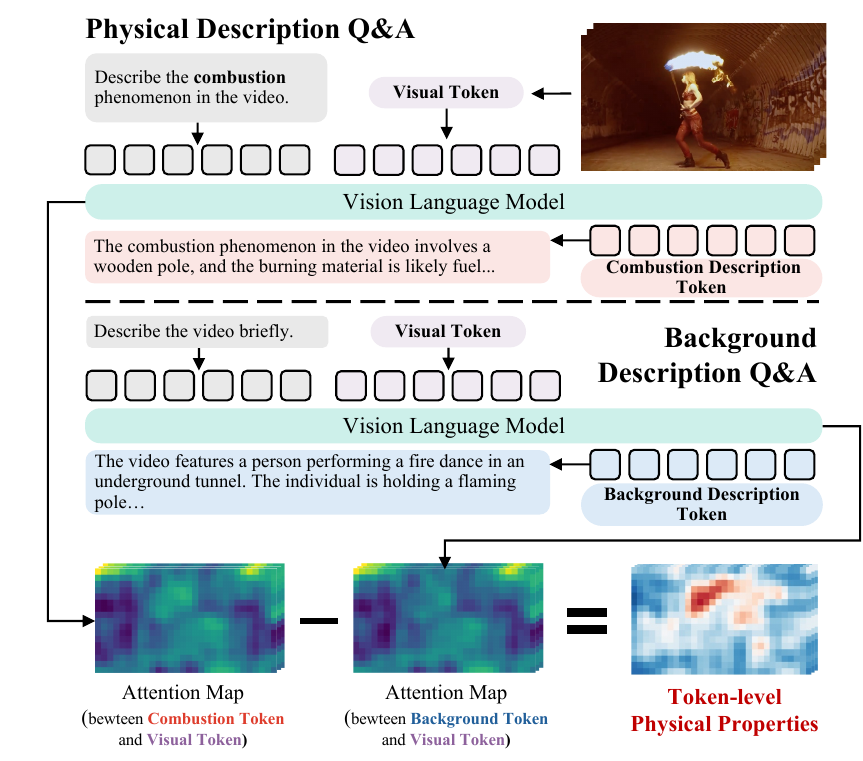}
    \vspace{-12pt}
    \caption{Pipeline for annotating token-level physical attributes using a VLM.}
    \vspace{-12pt}
    \label{fig:token}
\end{figure}



\begin{figure}
    \centering
    \includegraphics[width=\columnwidth]{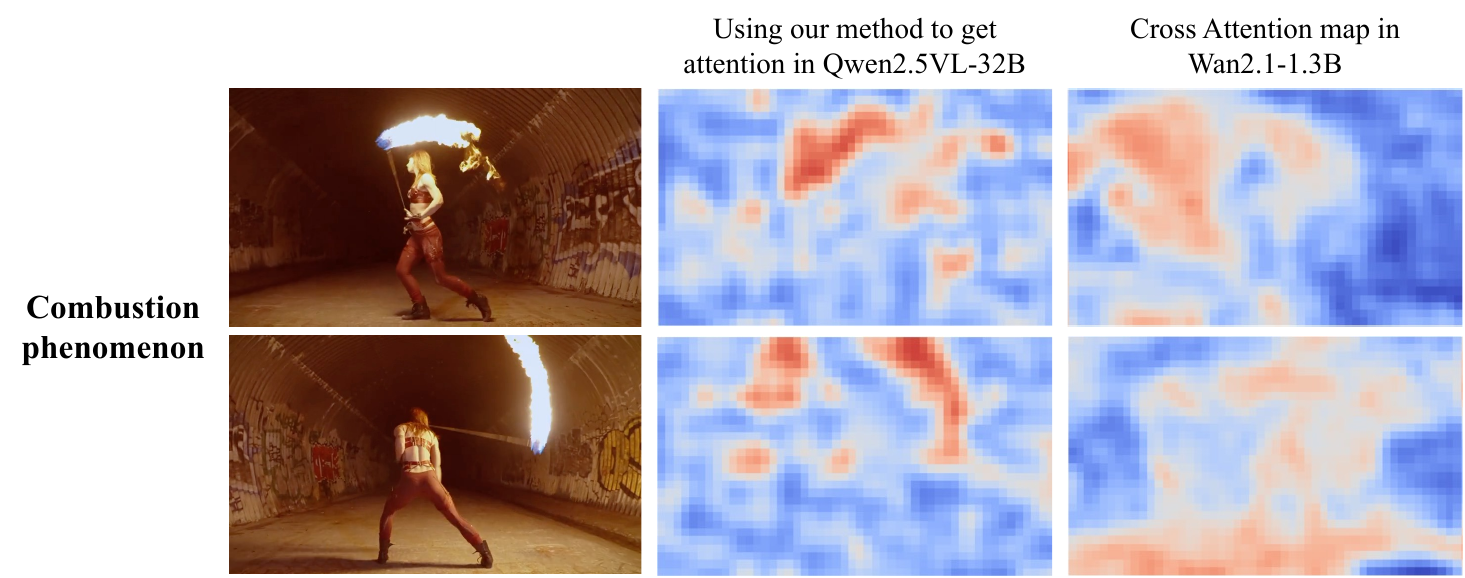}
    \vspace{-12pt}
    \caption{Study of the attention localization capabilities of VDM and VLM. The VDM cross-attention maps are obtained by adding 10\% noise and then denoising. As shown, despite minor imperfections, the VLM-based approach more accurately identifies the locations of the corresponding physical phenomena.}
    \vspace{-12pt}
    \label{fig:vdmvlm}
\end{figure}


To enable token-level physical priors to respond anisotropically to physical laws, the Refinement Router must infer the physical attributes of each token and activate the appropriate refinement experts. We find that VLMs exhibit stronger spatial understanding of physical dynamics than generative models, as shown in Figure~\ref{fig:vdmvlm}. They can also perceive the physical properties of individual video tokens with higher accuracy. Based on this observation, we align the perception capability of the refinement router with the understanding provided by the VLM.

To obtain the supervision signal, we feed a question about a target physical phenomenon into the VLM along with the video, as illustrated in Figure~\ref{fig:token}. The prompt asks the VLM to describe the phenomenon in the video. From the generated text, we extract the corresponding video tokens as key tokens and the relevant text tokens as query tokens. We then retrieve the attention scores between these query and key tokens, giving a preliminary localization of where the phenomenon occurs. To reduce labeling noise, we query the VLM again using a short, generic prompt that avoids specific objects or events. Using the same query–key selection process, we obtain a background attention map. Subtracting this background map from the phenomenon map yields the final token-level alignment targets $\boldsymbol{Q}_r \in \mathbb{R}^{N \times E{\text{attn}}}$, where $E_{\text{attn}}$ is the number of physical laws used for refinement. This procedure improves both the accuracy and the sharpness of the high-attention regions.


We define a mask $\boldsymbol{M} \in \mathbb{R}^{N \times E_{\text{attn}}}$. The mask is constructed as follows: (1) For efficiency, we annotate only the physical phenomena likely present in the video and set entries for annotated categories $e_{\text{marked}}$ to 1 and others to 0. (2) Some values in $\boldsymbol{Q}_r$ are negative, indicating that the phenomenon is not prominent in those regions. We update the mask element-wise as $\boldsymbol{M} = \boldsymbol{M} \wedge \text{sign}(\boldsymbol{Q}_r)$ to drop those regions. The resulting mask $\boldsymbol{M}$ highlights high-salience regions for the annotated phenomena. The fine-grained alignment loss is
\begin{equation}
    \mathcal{L}_{\text{fine-align}} = \sum_{\boldsymbol{M}^{i, e} = 1}\|{\boldsymbol{P}'}_r^{i, e} - \boldsymbol{Q}_r^{i, e}\|_2,
\end{equation}
where $\boldsymbol{P}'_r = [{\boldsymbol{\rho}'}_{r}^{(1)}, {\boldsymbol{\rho}'}_{r}^{(2)}, \dots, {\boldsymbol{\rho}'}_{r}^{(N)}]^\top$. Here, $\boldsymbol{\rho}'_r$ is obtained by passing the refinement router output $\boldsymbol{\rho}r$ through an MLP that expands the dimension from $E_r$ to $E{\text{attn}}$. Beyond dimension matching, this MLP also reduces the training conflict~\cite{videorepa} that arises when alignment signals are applied directly. To further stabilize training and encourage specialization in the refinement experts, we introduce a standard load-balancing auxiliary loss~\cite{shazeer2017outrageously} $\mathcal{L}_{\text{fine-balance}}$ on the refinement router’s outputs.
As a result, the final training objective combines the three proposed losses with the standard diffusion loss, formulated as follows:
\begin{equation}
    \mathcal{L} = \mathcal{L}_{\text{diffusion}} + \lambda_1\mathcal{L}_{\text{coarse}} + \lambda_2\mathcal{L}_{\text{fine-align}} + \lambda_3\mathcal{L}_{\text{fine-balance}},
    \label{eq:7}
\end{equation}
where $\lambda_1, \lambda_2$ and $\lambda_3$ represent the weights, respectively.



%% file: sec/4_exp.tex
\section{Experiment}
\label{sec:exp}

\begin{table}
  \caption{Results on VideoPhy2 benchmark. The best results are highlighted in \textbf{bold}, and the second-best results are \underline{underlined}.}
  \label{tab:videophy2}
  \centering
  \setlength{\tabcolsep}{3.5pt}
  \begin{tabularx}{\columnwidth}{@{\extracolsep{\fill}}c|ccc|ccc}
    \toprule
    \multirow{2}{*}[-0.5ex]{Method} & \multicolumn{3}{c|}{ALL} & \multicolumn{3}{c}{HARD} \\
    \cmidrule(lr){2-4} \cmidrule(lr){5-7}
    & PC & SA & Joint & PC & SA & Joint \\
    \midrule
    HunyuanVideo \cite{hunyuan} & 64.2 & 19.2 & 24.7 & \underline{52.2} & 7.2 & 5.0 \\
    Cosmos \cite{cosmos} & 54.6 & 26.2 & 22.6 & 40.0 & 5.1 & 4.0 \\
    Cosmos-Predict2.5 & 61.3 & 27.7 & 23.1 & 50.6 & 8.3 & 5.6 \\
    Wan2.1-14B \cite{wan} & 60.5 & 29.0 & 24.7 & 45.6 & 8.3 & 3.9 \\
    \midrule
    Wan2.1-1.3B & 57.8 & 30.0 & 24.8 & 36.7 & \underline{11.7} & 5.6 \\
    \rowcolor{gray!10} + ProPhy & 65.0 & \underline{32.0} & \underline{26.5} & 48.9 & \textbf{12.2} & \textbf{7.2} \\
    \midrule
    CogVideoX-5B \cite{cogvideox} & 67.2 & 29.0 & 22.3 & 51.1 & 9.6 & 5.0 \\
    + WISA \cite{wisa} & \underline{69.1} & 31.5 & 25.8 & 51.7 & 11.1 & 5.0 \\
    + VideoREPA \cite{videorepa} & \textbf{72.5} & 24.2 & 22.0 & \underline{52.2} & 7.8 & 5.6 \\
    \rowcolor{gray!10} + ProPhy & \textbf{72.5} & \textbf{32.2} & \textbf{26.7} & \textbf{52.8} & \underline{11.7} & \underline{6.1} \\
    \bottomrule
  \end{tabularx}
  \vspace{-12pt}
\end{table}

\begin{table*}
  \caption{Results on VBench quality score. For each method, the best performance relative to its base model is highlighted in \textbf{bold}.}
  \label{tab:vbench}
  \centering
  \setlength{\tabcolsep}{3.5pt}
  \begin{tabularx}{\linewidth}{X|*{7}{c}|c}
  \toprule
  \multirow{2}{*}{Method} & Subject & Background & Temporal & Motion & Dynamic & Aesthetic & Imaging & \textbf{\ Quality\ } \\
  & Consistency & Consistency & flickering & Smoothness & Degree & Quality & Quality & \textbf{Score} \\ 
  \midrule
  CogVideoX-5B \cite{cogvideox} & \underline{92.9} & \textbf{95.6} & 97.9 & \underline{98.7} & 46.8 & \underline{48.2} & 51.1 & 76.8 \\
  + WISA \cite{wisa} & 91.7 & 94.9 & \underline{98.1} & 98.3 & \underline{49.7} & \textbf{49.4} & \underline{54.4} & \underline{78.8} \\
  + VideoREPA \cite{videorepa} & 91.3 & 94.5 & \textbf{98.3} & \textbf{99.0} & 38.7 & 47.7 & 48.5 & 77.0 \\
  \rowcolor{gray!10} + ProPhy & \textbf{93.2} & \underline{95.1} & \textbf{98.3} & \underline{98.7} & \textbf{72.0} & 47.6 & \textbf{66.0} & \textbf{81.0} \\
  \midrule
  Wan2.1-1.3B \cite{wan} & 88.8 & 92.8 & 96.7 & 97.9 & 71.3 & 47.7 & 57.3 & 77.3 \\
  \rowcolor{gray!10} + ProPhy & \textbf{89.7} & \textbf{93.1} & \textbf{97.1} & \textbf{99.0} & \textbf{78.8} & \textbf{47.8} & \textbf{58.4} & \textbf{79.0} \\
  \bottomrule
  \end{tabularx}
\end{table*}


\subsection{Implementation Details}

\paragraph{Training}

We build ProPhy upon two open-source video diffusion models, Wan2.1-1.3B~\cite{wan} and CogVideoX-5B~\cite{cogvideox}. These models are fine-tuned with our ProPhy framework to validate its effectiveness. We randomly sample 20K videos from the WISA-80K dataset~\cite{wisa} as the training data, and employ Qwen2.5-VL-32B~\cite{bai2025qwen2} to obtain token-level physical annotations. For the hybrid loss in Equation~\ref{eq:7}, we use fixed coefficients $\lambda_1 = 0.1$, $\lambda_2 = 0.02$, and $\lambda_3 = 0.01$. This configuration works reliably across both backbones without additional tuning. Further implementation details are provided in the \textbf{Supplementary Material}.
\vspace{-12pt}
\paragraph{Evaluation}
We evaluate our generated videos using the \textbf{VideoPhy2} benchmark \cite{bansalVideoPhy2ChallengingActionCentric2025}. VideoPhy2 is an action-centric evaluation suite that measures video quality through \textit{physical commonsense} (PC) and \textit{semantic adherence} (SA), together with their joint pass rate (Joint). The benchmark includes 600 carefully curated prompts, along with a subset of 180 more challenging ones. For both PC and SA, VideoPhy2 assigns integer scores from 1 to 5. Following the official definition, we treat scores of 4 or higher as PC = 1 or SA = 1. A joint pass (Joint = 1) is achieved only when both PC = 1 and SA = 1. This joint rate is the primary metric for assessing the physical plausibility of generated videos. To show that ProPhy improves physical reasoning without sacrificing visual quality, we also assess our method on the quality‐oriented dimensions of \textbf{VBench}~\cite{huangVBenchComprehensiveBenchmark2023a}, using the identical set of 600 prompts for consistency.

\subsection{Quantitative Comparisons}

To validate the effectiveness of ProPhy, we compare it against its base models, Wan2.1~\cite{wan} and CogVideoX~\cite{cogvideox}. We further include several strong open-source T2V models (HunyuanVideo~\cite{hunyuan}, Cosmos~\cite{cosmos}, Cosmos-Predict2.5~\cite{cosmos}) as well as physics-enhanced approaches WISA~\cite{wisa} and VideoREPA~\cite{videorepa}. As shown in Table~\ref{tab:videophy2}, applying ProPhy to CogVideoX yields either the best or the second-best results across all metrics. On the entire VideoPhy2~\cite{bansalVideoPhy2ChallengingActionCentric2025} benchmark (\textit{ALL}), especially in terms of the \textit{Joint} metric, ProPhy delivers a substantial +19.7\% improvement on the CogVideoX, with consistent gains in PC and SA. On the more challenging \textit{HARD} subset, our progressive alignment framework also produces videos that exhibit stronger semantic consistency and more faithful physical behaviors.

As presented in Table~\ref{tab:vbench}, ProPhy also maintains strong generation capability on VBench metrics. Our improvements are most notable in the \textit{Dynamic Degree} dimension, confirming that our progressive physical alignment framework enhances the model’s ability to capture highly dynamic physical behaviors. In addition, the aggregated \textit{Quality Score}, computed as a weighted combination of the seven VBench dimensions~\cite{huangVBenchComprehensiveBenchmark2023a}, shows that ProPhy produces videos of overall higher quality.

\subsection{Qualitative comparison}

\begin{figure*}
    \centering
    \includegraphics[width=\linewidth]{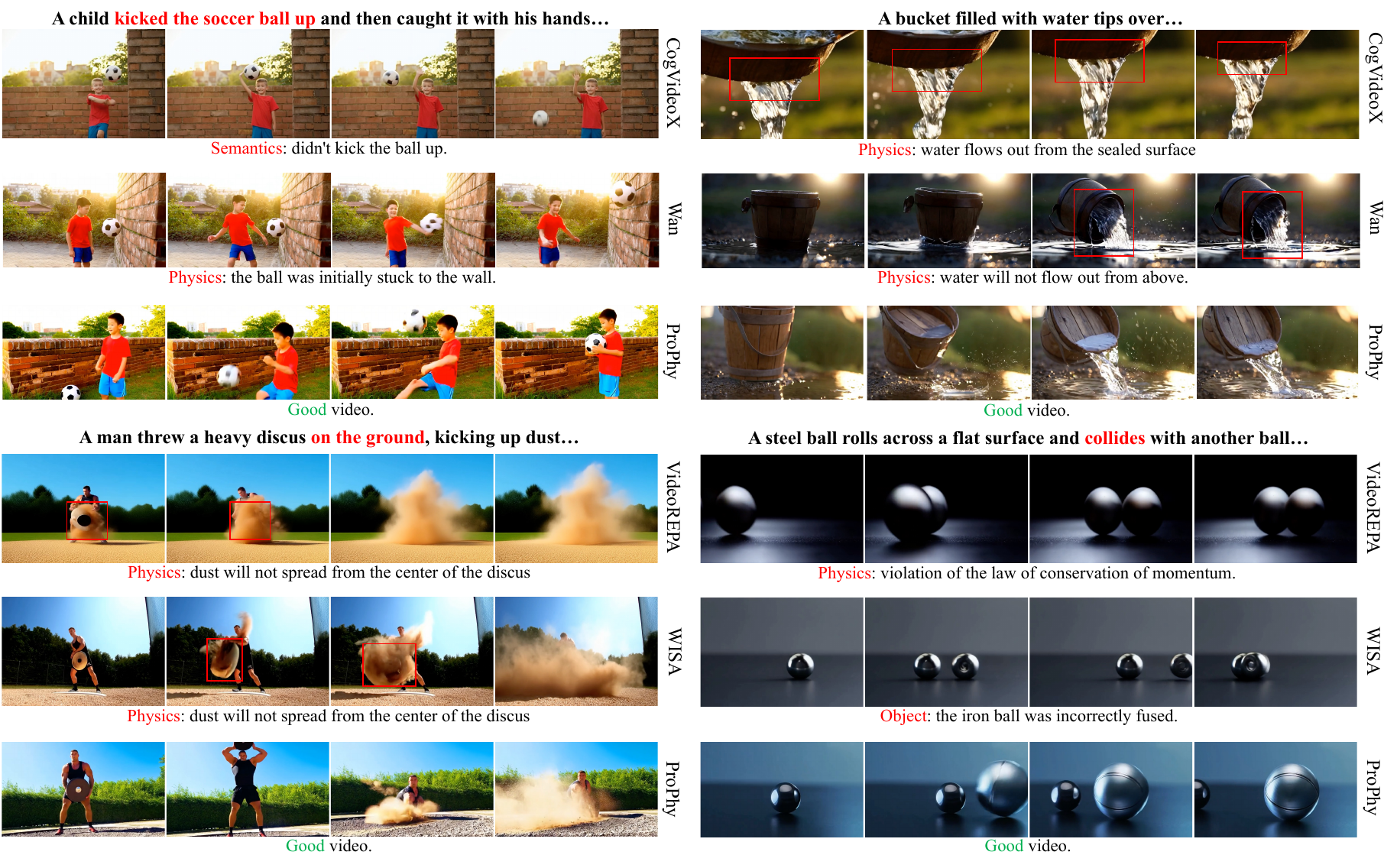}
    \vspace{-12pt}
    \caption{Qualitative comparison among ProPhy, CogVideoX, Wan2.1, and existing physics-aware methods.}
    \vspace{-12pt}
    \label{fig:qual}
\end{figure*}

As shown in Figure~\ref{fig:qual}, we present a qualitative comparison between ProPhy, the base video generators (CogVideoX~\cite{cogvideox} and Wan2.1~\cite{wan}), and state-of-the-art physics-aware approaches (VideoREPA~\cite{videorepa} and WISA~\cite{wisa}). In the discus-throw scenario, existing models lack fine-grained physical alignment, often coupling the dust plume with the discus trajectory. In contrast, ProPhy correctly triggers dust emission only when the discus contacts the ground. In the iron-ball collision scenario, prior methods violate momentum conservation, exhibiting penetration artifacts and incorrect spatial reasoning. ProPhy, however, produces collisions that obey momentum conservation: after the larger ball strikes the smaller one, kinetic energy is transferred, causing the smaller ball to start moving from rest.

\subsection{Ablation Study}

\begin{table}
  \caption{Ablation study results on ProPhy with Wan2.1-1.3B as the base model. \textit{LoRA} indicates that the Physical Branch is removed, and LoRA is applied to the backbone.}
  \label{tab:ablation}
  \centering
  \begin{tabularx}{\columnwidth}{@{\extracolsep{\fill}}ccc|ccc}
    \toprule
    \multicolumn{3}{c|}{Settings} & \multirow{2}{*}[-0.5ex]{PC} & \multirow{2}{*}[-0.5ex]{SA} & \multirow{2}{*}[-0.5ex]{Joint} \\
    \cmidrule(lr){1-3}
    PB & SEB & REB & & & \\
    \midrule
    \multicolumn{3}{c|}{Baseline} & 57.8 & 30.0 & 24.8 \\
    \textit{LoRA} & - & - & 58.2 & 30.8 & 24.8 \\
    \textit{LoRA} & - & \checkmark & 62.7 & 31.2 & 25.5 \\
    \textit{LoRA} & \checkmark & - & 62.2 & 30.8 & 25.2 \\
    \textit{LoRA} & \checkmark & \checkmark & \underline{64.0} & 31.2 & 26.0 \\
    \checkmark & - & - & 58.7 & 30.0 & 25.7 \\
    \checkmark & - & \checkmark & 63.7 & \textbf{32.2} & \underline{26.2} \\
    \checkmark & \checkmark & - & 63.5 & 31.5 & 26.0 \\
    \checkmark & \checkmark & \checkmark & \textbf{65.0} & \underline{32.0} & \textbf{26.5} \\
    \bottomrule
  \end{tabularx}
\end{table}

\begin{table}
  \caption{Ablation study on the roles of relative loss and absolute loss during training.}
  \label{tab:loss}
  \centering
  \begin{tabularx}{\columnwidth}{@{\extracolsep{\fill}}l|ccc}
      \toprule
      Settings & PC & SA & Joint \\
      \midrule
      baseline & 57.8 & 30.0 & 24.8 \\
      BCE loss in SEB & \underline{64.3} & \textbf{32.0} & \underline{26.3} \\
      only align loss in REB & 58.3 & 26.5 & 21.6 \\
      only load balance loss in REB & 64.0 & 31.7 & \underline{26.3} \\
      ProPhy & \textbf{65.0} & \textbf{32.0} & \textbf{26.5} \\
      \bottomrule
  \end{tabularx}
\end{table}

To validate the effectiveness of our progressive alignment framework, we use Wan2.1-1.3B as the baseline model. We conduct extensive ablation studies on this backbone and evaluate all variants on the VideoPhy2~\cite{bansalVideoPhy2ChallengingActionCentric2025} benchmark. As shown in Table~\ref{tab:ablation}, we compare a LoRA fine-tuning of the backbone (without the Physical Branch) with our Physics-Branch conditioning. Under the same number of training steps, the Physical Branch provides larger gains than plain LoRA. In addition, incorporating both the SEB and the REB further improves physical commonsense and semantic adherence.

As shown in Table~\ref{tab:loss}, we ablate the relative loss in the SEB and the two losses in the REB. In the SEB, replacing our relative-distance loss with BCE improves SA but weakens PC and Joint, indicating that the relative formulation provides more effective guidance. In the REB, using only the absolute alignment loss causes clear degradation, while using only the load-balance loss improves SA but reduces PC due to missing fine-grained cues. The full combination in ProPhy achieves the best performance across all metrics, confirming the effectiveness of our alignment design. Additional ablation studies are provided in the \textbf{Supplementary Material}.

\subsection{Physics Learned by Experts}

\begin{figure}
    \centering
    \includegraphics[width=\columnwidth]{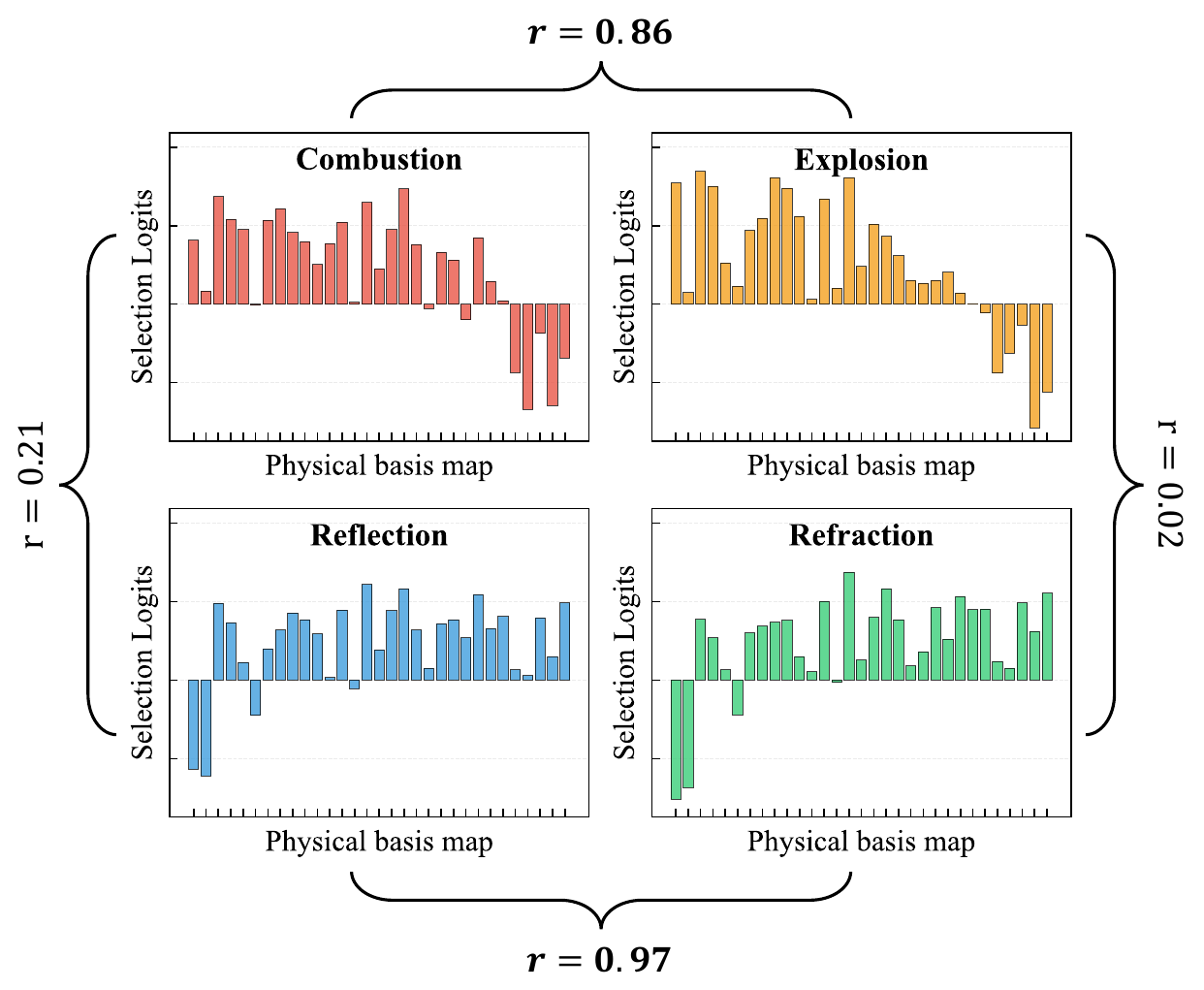}
    \vspace{-12pt}
    \caption{Analysis of the semantic router. $r$ represents the Pearson correlation coefficient calculated between different distributions.}
    \vspace{-12pt}
    \label{fig:semantic}
\end{figure}

\begin{figure}
    \centering
    \includegraphics[width=\columnwidth]{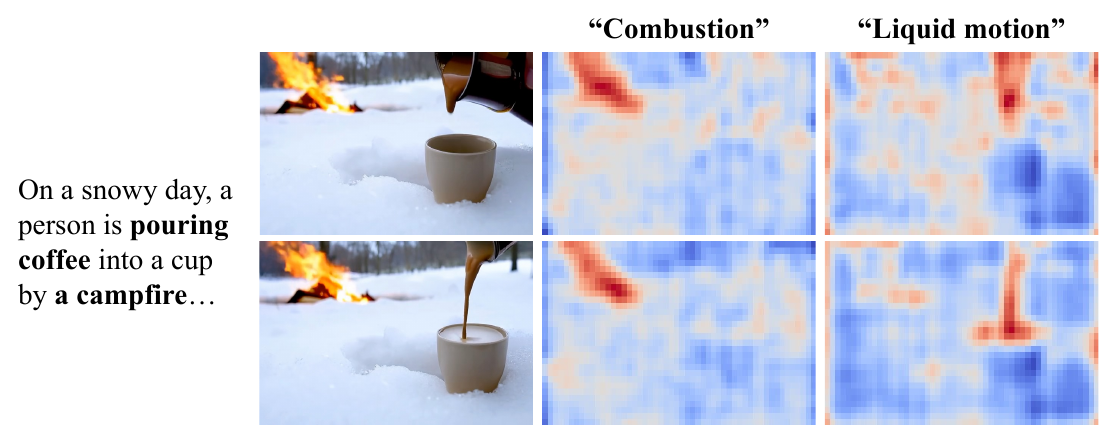}
    \vspace{-12pt}
    \caption{Refinement router expert maps. High-activation regions accurately localize where corresponding physical events occur, demonstrating the REB’s fine-grained physical alignment.}
    \vspace{-12pt}
    \label{fig:refinement}
\end{figure}

\begin{figure}
    \centering
    \includegraphics[width=\columnwidth]{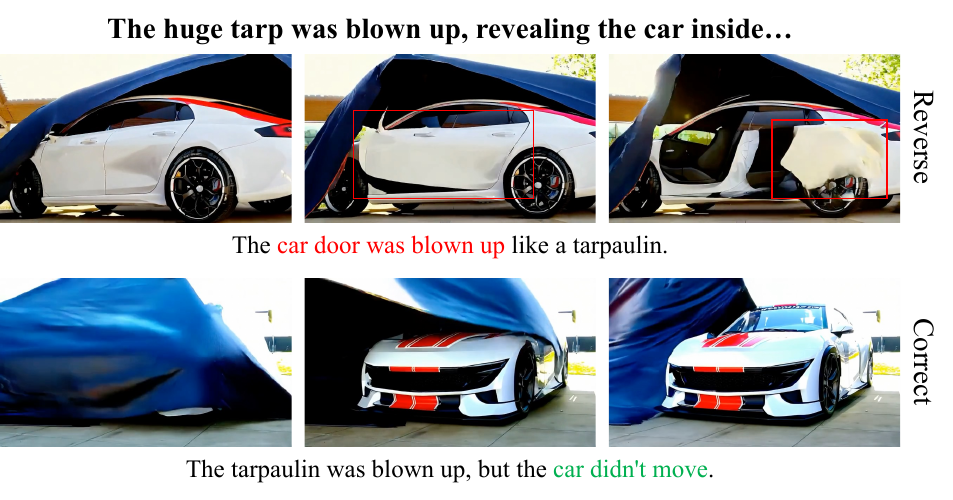}
    \vspace{-12pt}
    \caption{Physical attribute transfer via expert inversion. Flipping semantic-router logits injects incorrect physical cues, causing implausible behaviors (\textit{e.g., a rigid car door fluttering}), revealing that different experts encode distinct physical priors.}
    \vspace{-12pt}
    \label{fig:invert}
\end{figure}

To analyze how physical knowledge is captured in both the SEB and REB, we design the following experiments.

For the SEB, we use WISA~\cite{wisa} labels to group prompts by physical category and randomly sample 100 prompts per category. These prompts are fed into the semantic router, and we examine the resulting logits. As shown in Figure~\ref{fig:semantic}, the router exhibits meaningful structure: physically related categories (e.g., combustion and explosion) show high Pearson correlation, whereas unrelated ones (e.g., explosion vs. refraction) show low similarity. This indicates that the SEB learns coherent physical semantics.

For the REB, we visualize the projected logits of the refinement router. Figure~\ref{fig:refinement} shows that regions with high activation reliably align with where corresponding physical events occur in the video, showing that the REB performs fine-grained physical alignment that guides the generator.

Finally, we examine the attribution of physical priors within ProPhy's MoE~\cite{moe} structure. During inference, we manually manipulate the refinement router’s logits to assign ``incorrect'' physical experts to specific objects. As shown in Figure~\ref{fig:invert}, this leads to counter-intuitive physical behaviors, such as a rigid car door exhibiting cloth-like deformation. These results demonstrate that individual experts indeed capture distinct physical priors, suggesting that ProPhy’s internal representations are physically-grounded and possess the potential for more explicit attribute control in future work.

%% file: sec/5_conc.tex
\section{Conclusion}
\label{sec:conc}
In this paper, we presented ProPhy, a Progressive Physical Alignment Framework for physics-aware video generation. We first identified the limitations of existing models in maintaining physical consistency, particularly their implicit physical guidance and lack of fine-grained spatial alignment. To address these issues, we introduced a two-stage MoPE mechanism to extract hierarchical physical priors, along with a physical alignment strategy that transfers physical reasoning capabilities from VLMs into the generation process. Together, these designs enable the model to capture fine-grained, anisotropic physical dynamics and produce physically coherent videos. Extensive experiments demonstrate that ProPhy achieves state-of-the-art performance, while also validating the importance of fine-grained alignment in producing physically plausible video content.


\paragraph{Limitations} The regions annotated for physical phenomena inevitably contain noise, and simple region-level physical categorization captures only coarse surface patterns. Future work may integrate these aligned regions with the governing physical differential equations to inject more interpretable and principled physical knowledge into video generation models.